\documentclass[letterpaper, 10 pt, conference]{ieeeconf}  % Comment this line out if you need a4paper   

\IEEEoverridecommandlockouts                              % This command is only needed if 
\usepackage{graphics} % for pdf, bitmapped graphics files
\usepackage{epsfig} % for postscript graphics files
\usepackage{mathptmx} % assumes new font selection scheme installed
\usepackage{times} % assumes new font selection scheme installed
\usepackage{amsmath} % assumes amsmath package installed
\usepackage{amssymb}  % assumes amsmath package installed

\usepackage{bbding}

\usepackage[utf8]{inputenc} % allow utf-8 input
\usepackage[T1]{fontenc}    % use 8-bit T1 fonts
\usepackage[pagebackref=true,breaklinks=true,letterpaper=true,colorlinks,bookmarks=true]{hyperref}
\usepackage{url}            % simple URL typesetting
\usepackage{booktabs}       % professional-quality tables
\usepackage{amsfonts}       % blackboard math symbols
\usepackage{nicefrac}       % compact symbols for 1/2, etc.
\usepackage{microtype}      % microtypography
\usepackage{color}
\usepackage[table]{xcolor}
\usepackage{cleveref}
\usepackage{graphicx}
\usepackage{float}
\usepackage{multirow}
\usepackage{diagbox}

\usepackage{algorithm}
\usepackage{algpseudocode}
\usepackage{arydshln}
\usepackage{colortbl}
\usepackage{multicol}
\usepackage{multirow}
\usepackage{listings}
\definecolor{dkgreen}{rgb}{0,0.6,0}
\definecolor{gray}{rgb}{0.5,0.5,0.5}
\definecolor{mauve}{rgb}{0.58,0,0.82}
\title{\LARGE \bf
RenderOcc: Vision-Centric 3D Occupancy Prediction \\ with 2D Rendering Supervision
}

\author{Mingjie Pan$^{1,2*}$, Jiaming Liu$^{1*}$, Renrui Zhang$^{3*}$,  
Peixiang Huang$^{2}$,   \\
Xiaoqi Li$^{1}$, 
Hongwei Xie$^{4}$, Bing Wang$^{5}$, 
Li Liu$^{2\dagger}$, Shanghang Zhang$^{1\dagger}$
\thanks{$^1$ Mingjie Pan, Jiaming Liu, Xiaoqi Li and Shanghang Zhang are with National Key Laboratory for Multimedia Information Processing, School of CS, Peking University. 
 $^2$ Mingjie Pan, Peixiang Huang and Li Liu are with Xiaomi Car. $^3$ Renrui Zhang is with CUHK MMLAB. 
 $^4$ Hongwei Xie is with Nanjing University. 
 $^5$ Bing Wang is with Nanyang Technological University.
 }
\thanks{$*$ The first three authors contributed equally. }
\thanks{$\dagger$ Corresponding to shanghang@pku.edu.cn or liuli.ll9412@gmail.com.}
}

\begin{document}
\maketitle

\begin{abstract}
3D occupancy prediction holds significant promise in the fields of robot perception and autonomous driving, which quantifies 3D scenes into grid cells with semantic labels.
Recent works mainly utilize complete occupancy labels in 3D voxel space for supervision. However, the expensive annotation process and sometimes ambiguous labels have severely constrained the usability and scalability of 3D occupancy models.
To address this, we present RenderOcc, a novel paradigm for training 3D occupancy models only using 2D labels. 
Specifically, we extract a NeRF-style 3D volume representation from multi-view images, and employ volume rendering techniques to establish 2D renderings, thus enabling direct 3D supervision from 2D semantics and depth labels. 
Additionally, we introduce an Auxiliary Ray method to tackle the issue of sparse viewpoints in autonomous driving scenarios, which leverages sequential frames to construct comprehensive 2D rendering for each object.
To our best knowledge, RenderOcc is the first attempt to train multi-view 3D occupancy models only using 2D labels, reducing the dependence on costly 3D occupancy annotations. 
Extensive experiments demonstrate that RenderOcc achieves comparable performance to models fully supervised with 3D labels, underscoring the significance of this approach in real-world applications. 
Our code is available at \href{https://github.com/pmj110119/RenderOcc}{https://github.com/pmj110119/RenderOcc}.

\end{abstract}

\section{Introduction}
% 3D感知是自动驾驶等基于视觉的机器人系统的重要组成部分，目前流行的视觉感知任务之一是3D 语义Occupancy Prediction，它将3D空间量化为带语义标签的网格单元。相比3D box，3D occupancy是几何感知的，能够精细的描述出不同的物体的背景的形状，而不仅仅是用一个box指示粗粒度的位置和大小。3D Occupancy已经在工业界和学术界被广泛advocated，在场景理解和各种下游任务上展现出了潜力\cite{occnet}.
% 3D perception plays an important role in vision-based robotic systems, particularly in autonomous driving. One of the currently popular tasks in visual perception is 3D Semantic Occupancy Prediction, which quantifies 3D space into grid cells with semantic labels. Unlike the 3D bounding box, 3D occupancy provides geometric perception, allowing for a detailed description of the shape of different objects' surroundings, rather than just indicating coarse-grained position and size with a box. This approach has gained widespread recognition and adoption in both the industry and academic.

% Renrui
Perceiving the 3D world plays an important role in vision-based robotic systems and autonomous driving \cite{arnold2019survey, hu2023planning}. One of the currently popular tasks in 3D vision is semantic occupancy prediction. It requires quantifying continuous 3D space into grid cells and predicting semantic labels for each voxel. Compared to 3D object detection \cite{philion2020lift, wang2022detr3d}, 3D occupancy provides more fine-grained geometric perception, allowing for a detailed understanding of object shapes and scene-level geometries, rather than the coarse-grained 3D bounding box. To this point, many solutions have been proposed and gained widespread adoption in both industry and academia~\cite{tesla,occnet}.

Existing methods mostly rely on complete 3D occupancy labels for supervision. However, directly annotating 3D occupancy is extremely challenging and expensive. For one thing, after creating $\sim 30000$ frames of 3D occupancy labels based on other pre-annotated 3D labels, it still requires costly $\sim 4000$ human hours for purification \cite{openoccupancy}.
For another, the complexity of 3D space would cause ambiguous labels, distracting the training of occupancy models. 
We conducted a comparison on several benchmarks and found that even when they use identical raw data, the occupancy labels they produce still exhibit a 10-15\% difference 
% \textcolor{red}{[TO BE CHECK]} 
\cite{openoccupancy, occ3d, surroundocc}. This reflects the inherent ambiguity of occupancy annotation, which further restricts the practical application of 3D occupancy tasks in real-world scenarios.

\begin{figure}[t]
\includegraphics[width=1.0\linewidth]{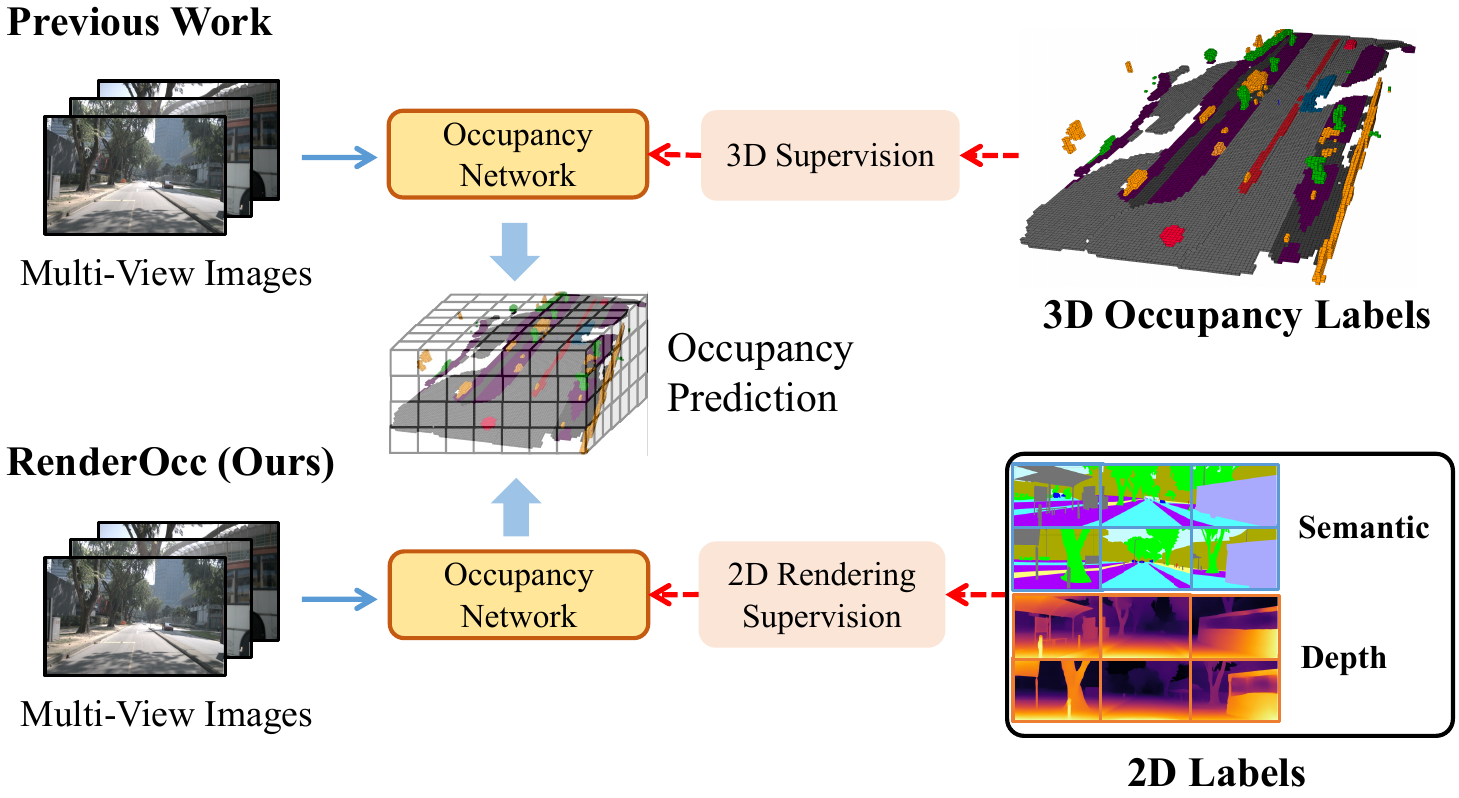}
\centering
\vspace{-0.5cm}
\caption{
% [小标题]？？？
\textbf{RenderOcc represents a new training paradigm.}  
Unlike previous works that focus on supervising with costly 3D occupancy labels, our proposed RenderOcc utilizes 2D labels to train the 3D occupancy network. Through 2D rendering supervision, the model benefits from fine-grained 2D pixel-level semantic and depth supervision.
}
\label{fig:intro}
\vspace{-0.5cm}
\end{figure}

Towards these issues aforementioned, we introduce RenderOcc, a novel paradigm for training 3D occupancy models using 2D labels, free from any 3D-space annotations. As shown in Fig. \ref{fig:intro}, the goal of RenderOcc is to eliminate the dependency on 3D occupancy labels and rely solely on pixel-level 2D semantics for network supervision during training. 
% Specifically, RenderOcc utilizes multi-view images to construct a 3D volume representation in the style of NeRF \cite{nerf}. Volume rendering is then applied, facilitating direct supervision using 2D semantic and depth labels.
% RenderOcc derives a NeRF-style 3D volume representation from multi-view images and employs volume rendering to enable direct supervision from 2D semantic and depth labels.
In particular, it constructs a NeRF-style 3D volume representation from multi-view images and utilizes advanced volume rendering techniques to generate 2D renderings. This approach enables us to provide direct 3D supervision using only 2D semantics and depth labels.
With such a 2D rendering supervision, the model learns multi-view consistency by analyzing intersecting frustum rays from various cameras, obtaining a deeper understanding of geometric relationships in 3D space. 
Importantly, it is worth noting that autonomous driving scenarios often involve limited viewpoints, which can hinder the effectiveness of rendering supervision. 
Considering this, we introduce the concept of Auxiliary-Ray, which leverages rays from adjacent frames to enhance the multi-view consistency constraints for the current frame. 
% This enables more effective supervision of voxel features. 
% Furthermore, we have designed a dynamic sampling training strategy for auxiliary rays, aimed at mitigating the additional training costs associated with these rays, thereby achieving efficient and beneficial training.
Moreover, we have developed a dynamic sampling training strategy for auxiliary rays, which not only screens out misaligned rays, but also simultaneously mitigates the additional training costs associated with them.

To evaluate the effectiveness of our proposed method, we conduct extensive experiments on two widely recognized benchmarks, NuScenes \cite{nuscenes} and SemanticKiTTI \cite{semantickitti}. Remarkably, with only 2D supervision, our method achieved competitive performance compared with models supervised by 3D labels. This highlights the considerable potential of our approach in real-world perception systems. Meanwhile, our method outperforms previous state-of-the-art 3D occupancy prediction methods by leveraging both 2D labels and corresponding 3D labels.
The main contributions are summarized as follows:

% 我们introduce了RenderOcc，一种基于2D rendering supervision的3D occupancy训练范式。我们首次尝试了仅利用2D标签训练多视角3D Occupancy网络，探索了一条不依赖于昂贵且制作困难的3D标签的路线。
\textbf{1)} We introduce RenderOcc, a 3D occupancy framework based on 2D rendering supervision. We make the first attempt to train multi-view 3D occupancy networks solely using 2D labels, discarding the costly and challenging 3D annotation.

% For the first time, we incorporate 2D supervision into multi-view vision occupancy prediction, exploring a path that does not rely on expensive and intricate 3D labels.

% 我们引入了Auxiliary Rays来解决自动驾驶场景视角稀疏的问题。并设计了Weighted Ray Sampling (WRS) 策略对射线进行Balance和Purification。我们的工作大大提高了2D rendering supervision的收敛性。
\textbf{2)} 
To learn a favorable 3D voxel representation from limited viewpoints,
% In order to construct comprehensive 3D voxel representation, 
we introduce Auxiliary-Rays to tackle the challenge of sparse viewpoints in autonomous driving scenarios. 
Meanwhile, we design a dynamic sampling training strategy for balancing and purifying auxiliary rays.
% auxiliary rays with the dual purpose of screening out misaligned rays
% To achieve efficient training, we also designed a Weighted-Ray-Sampling strategy for balancing and purifying auxiliary rays. 
% Our work significantly enhances the convergence of 2D rendering supervision.

% In RenderOcc, we introduce Auxiliary Rays to complement supervision information, enabling the model to better benefit from multi-view consistency constraints. Additionally, we have designed the Weighted Ray Sampling (WRS) strategy to balance and purify auxiliary rays, achieving superior performance with significantly reduced training costs.

% extensive的实验show了RenderOcc能够在仅使用2D标签的情况下取得与使用3D标签的Baselines相比有竞争力的性能。这证明了2D图像监督用于3D occupancy训练的可行性和前景。
\textbf{3)} Extensive experiments show that RenderOcc achieves competitive performance when using only 2D labels, compared to baselines that supervised by 3D labels. This showcases the feasibility and potential of 2D image supervision for 3D occupancy training.

\section{Related work}
\noindent\textbf{3D Object Detection.} 
3D object detection is a classic perception task in the fields of robot perception and autonomous driving \cite{pointpillars, pointnet, second, centerpoint, fcos3d, pointnet++}. Recently, vision-based 3D object detection has gained increased attention due to its low cost and rich semantic content, \cite{wang2022detr3d, bevdet,bevformer,bevdepth, bevstereo, bev-lgkd, petr, petrv2, bevsan, solofusion, sparse4d} realize efficient transformation from multiple perspective views to a unified 3D space in a single frame, achieving cross-view 3D perception.

\noindent\textbf{3D Occupancy Prediction.} 
3D occupancy prediction, which can generate a dense 3D voxelized semantic representation of a scene, is an ideal capability for autonomous vehicles. The release of the SemanticKITTI dataset has drawn attention to 3D occupancy prediction \cite{monoscene, lmscnet, aicnet, voxformer, tpvformer, occdepth}, but it lacks diversity in driving scenes and only evaluates front-view predictions \cite{semantickitti}.
Recent work has extended 3D occupancy prediction to multi-view surrounding scenes and produce large-scale benchmarks, which have significantly propelled the development of the field \cite{occ3d, openoccupancy, surroundocc, occnet}. Due to the difficulty of manually annotating dense 3D occupancy, existing work relies on extra 3D labels such as LiDAR segmentation to produce 3D occupancy labels Recent works \cite{simpleoccupancy, behind-the-scenes} discussed occupancy estimation based on rendering, but they do not account for semantic predictions.

% 此外，nerf也被探索用于从3D场景中学习高级语义信息，\cite{semantic-nerf}使用MLP预测3d point的语义类别和密度，\cite{semantic-ray} 进一步利用多视角重投影关系，让nerf能够推断出unseen scene的语义。这些工作展现出了nerf在高级语义任务中的巨大潜力。

%然而，基于MLP隐式建模的训练和推理时间都很长，最近的许多工作使用explicit representations，将NeRF的训练时间大大减少from hours to minutes per scene \cite{plenoxels, dvgo, instant-nerf, tensorfch}. 其中，\cite{plenoxels, dvgo}均使用dense grid representation来进行建模，直接在voxel中存储density和视图颜色相关的球谐系数/特征，\cite{instant-nerf}使用哈希表建模密度和颜色，\cite{tensorf}则通过张量分解减少了dense grid representation的内存占用，直接对low-rank components进行建模。

% 事实上，3d occupancy 表示同样是dense grid representation，和显式建模的NeRF存在许多共通之处。我们认为，NeRF仅通过多视角2D图像即可完成对3D scene的理解和建模，而现有的3d occupancy作为一种类似的representation，同样能够摆脱昂贵的3d标签、从图像本身中挖掘信息。

% 除了语义NeRF之外，
\noindent\textbf{3D Reconstruction and Rendering}
% 从2D图像推断对象或场景的3D几何是一项具有挑战性的任务，最近流行的方法通过神经辐射场对3D场景进行建模，并基于multi-view 2D images使用volumn rendering进行监督\cite{nerf,nerf++,mipnerf,mipnerf360}。为了提高训练效率，\cite{dvgo, plevoxels, instant-nerf, tensorf}进一步使用了voxel-based explicit representations取得显著的效果。与3D Occupancy Prediction不同的是，这类方法更关注于渲染质量，不关注语义理解且不具备泛化性。但是它们的训练思想对3D occupancy存在启发性。
Inferring the 3D geometry of objects or scenes from 2D images is a challenging task. Recent popular methods model 3D scenes through neural radiance fields and supervise them using volume rendering based on multi-view 2D images \cite{nerf,nerf++,mipnerf,mipnerf360}. To improve training efficiency, significant results have been achieved by further using voxel-based explicit representations \cite{dvgo, plenoxels, instant-nerf, tensorf}. Unlike 3D Occupancy Prediction, these methods focus on rendering quality, pay less attention to semantic understanding, and lack generalization. However, their training ideas can be enlightening for 3D occupancy.

\begin{figure*}[t]
\includegraphics[width=1.0\textwidth]{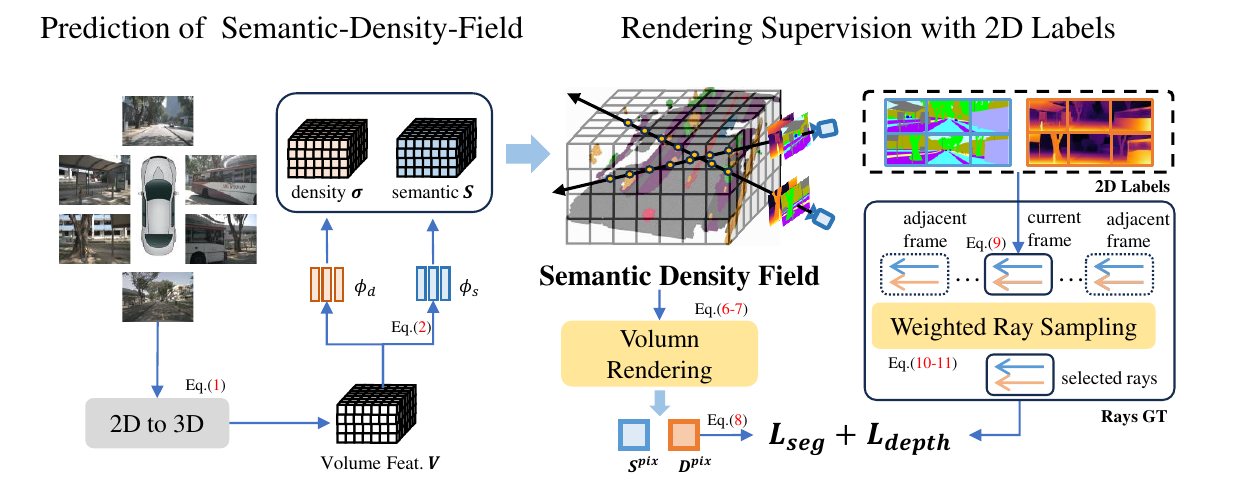}
\centering
\vspace{-0.4cm}
\caption{
\textbf{Overall framework of RenderOcc.} 
% 我们通过一个2D-to-3D network提取volumn features并为每个voxel预测density $\sigma$ and semantic $S$。由此我们生成了the Semantic Density Field (SDF), which can perform volumn rendering 以生成rendered的2D semantic and depth $\{S^{pix},D^{pix}\}$. 接着，我们用生成自2D图像标签的rays GT来和{S^{pix},D^{pix}\}$计算损失，通过rendering supervision with 2D labels训练3D occupancy网络。对于Rays GT的生成，我们从adjacent frames中提取辅助射线，并通过proposed Weighted Ray Sampling进行纯化得到最终用于监督的rays GT。
We extract volume features $V$ and predict density $\sigma$ and semantic $S$ for each voxel through a 2D-to-3D network. As a result, we generate the Semantic Density Field, which can perform volume rendering to generate rendered 2D semantics and depth $\{S^{pix},D^{pix}\}$. For the generation of Rays GT, we extract auxiliary rays from adjacent frames to supplement the rays of the current frame, and purify them using the proposed Weighted Ray Sampling strategy. Then, we calculate the loss with rays GT and $\{S^{pix},D^{pix}\}$, achieving rendering supervision with 2D labels. 
}
\label{fig:method}
\vspace{-0.2cm}
\end{figure*}

%------------------------------------------------------------------------
\section{Methods}
\label{sec:setup}
% \subsection{Preliminary}
% \textbf{Problem Setup.} 
\subsection{Problem Setup} 
We aim to predict a dense semantic volume, termed as 3D occupancy, of surrounding scenes with multi-camera RGB images. Specifically, for the vehicle at timestamp $t$, we take $N$ images $\{I^1,I^2,\cdots I^N\}$ as input and predict the 3D occupancy $O \in \mathbb{R}^{H\times W\times D\times L}$ as output, where $H,W,D$ denote the resolution of the volume and $L$ denotes the number of categories (including empty). 
Formally, the 3D occupancy prediction can be formulated as
\begin{align}
\label{semantic_eq}
V = \mathbb{G}(I^1,I^2,\cdots,I^N), \quad O = \mathbb{F}(V), 
\end{align}
where $\mathbb{G}$ is an neural network that extracts 3D volume feature $V \in \mathbb{R}^{H\times W\times D\times C}$ from $N$-view images, where $C$ denotes the feature dimension. $\mathbb{F}$ is responsible for transforming $V$ into occupancy representation, for which previous works \cite{openoccupancy, occ3d} tend to use MLP to achieve per-voxel classification. Considering that all existing approaches require complete 3D occupancy labels to supervise the voxel-level classification, we design a new concept to implement $\mathbb{F}$ and supervise $\{\mathbb{G}, \mathbb{F}\}$ with only 2D pixel-level labels.

% TODO!!!!
% \textbf{Design rationale.} Motivated by by elucidating the procedure for SDF prediction from RGB images. Subsequently, in Sec. \ref{RenderSup}, we detail how 2D-level supervision is achieved through rendering, reducing the reliance on 3D labels. In Sec

\subsection{Overall Framework}

Our overall framework is shown in Fig. \ref{fig:method}.
In Sec. \ref{SDF}, we first extract 3D volume features $V$ from multi-view RGB images with a 2D-to-3D network $\mathbb{G}$. Note that our framework is insensitive to the implementation of $\mathbb{G}$ and can flexibly switch between various BEV/Occupancy encoders such as \cite{bevdet,bevformer,voxformer}. 
Next in Sec. \ref{RenderSup}, we predict a volume density $\sigma$ and semantic logits $S$ for each voxel to generate the semantic density field (SDF). Subsequently, we perform volume rendering from SDF and optimize the network with 2D labels. Finally in Sec. \ref{AuxRays}, we illustrate the Auxiliary-Ray
training strategy for volume rendering to address the sparse viewpoints issue in autonomous
driving scenarios.

% [TODO]RenderOcc is a universal method that is built upon existing occupancy prediction models to enable the utilization of 2D labels for supervision. It can be summarized in the following steps: (a) extracting volumn features from multi-view images, (b) predicting the Semantic Density Field (SDF) – a versatile volumn-based representation, and (c) employing RenderSup to generate 2D semantic/depth maps from SDF, facilitating supervision with 2D labels.

% In sec. \ref{SDF}，我们首先介绍了如何从RGB图像中预测SDF。然后，我们在sec. \ref{RenderSup}中介绍了如何用Rendering的方式实现2D-level的监督、以摆脱对3D注释的强依赖。在sec. {AuxRays}中我们引入了辅助射线，通过增强多视角一致性约束来提高RenderOcc的性能。最后在sec. {RayMask}中，我们还为RenderOcc设计了一种Ray Masking策略，以少得多的训练开销取得更优的性能。
% In Sec. \ref{SDF}, we begin by elucidating the procedure for SDF prediction from RGB images. Subsequently, in Sec. \ref{RenderSup}, we detail how 2D-level supervision is achieved through rendering, reducing the reliance on 3D labels. In Sec. \ref{AuxRays}, we introduce Auxiliary Rays to enhance the multi-view consistency constraint and improve the performance. Finally, in Sec. \ref{RayMask}, we introduce a Ray Masking strategy designed for RenderOcc, enabling superior performance with significantly reduced training overhead.

% We will detail the Semantic-Density-Field in Sec. \ref{SDF}, RenderSup in Sec. \ref{RenderSup} and Auxiliary-Ray training strategy in Sec. \ref{AuxRays}.

\subsection{Semantic Density Field}
\label{SDF}
% 现有的Occupancy网络从多视角图像中提取volumn features，并进行voxel-wise的分类。RenderOcc将volumn features进一步提升为一种多功能的representation——Semantic Density Field (SDF)，用volumn density $\sigma$ and semantic logits $s$来表示场景。
Existing 3D occupancy methods extract volume features $V$ from multi-view images and perform voxel-wise classification \cite{voxformer,tpvformer,surroundocc,occnet} to generate 3D semantic occupancy. 
To employ 2D pixel-level supervision, our RenderOcc innovatively transforms $V$ into a versatile representation termed the Semantic-Density-Field (SDF). Given a volume feature map $V \in \mathbb{R}^{H \times W \times D \times C}$, the SDF encodes the scene by two representations: volume density $\sigma \in \mathbb{R}^{H \times W \times D}$ and semantic logits $S \in \mathbb{R}^{H \times W \times D \times L}$. 
Specifically, we simply adopt two MLPs $\{\phi_d, \phi_s\}$ to construct the SDF, formulated as
\begin{align}   
\label{de_diffusion}   
\sigma = softplus({\phi_d}(V)); \quad S= {\phi_s}(V),
\end{align}
where $\sigma$ additionally employs softplus activation to ensure that density values do not become negative. 
% 得到SDF后，我们可以对其进行任意视角的语义渲染。并用2D分割标签进行监督优化。这个过程在Sec. \ref{RenderSup}中进行介绍。在模型推理时，SDF可以直接转化为3D Occupancy。
Based on SDF, we gain the capability to perform semantic rendering from any viewpoints and get 2D supervision for optimization during training, which will be explained in Sec. \ref{RenderSup}. 

After optimized, SDF can be directly converted to 3D occupancy results. 
We filter out occupied voxels with $\sigma$ and determine their semantic categories based on $S$. The process can be formalized as follows:
\begin{align}
\label{semantic_eq2}
O(x,y,z) = \begin{cases} 
argmax(S(x,y,z)), &  \sigma(x,y,z) \geq \tau \\
empty \ label, &  \sigma(x,y,z) < \tau 
\end{cases},
\end{align}
where $\tau$ serves as the threshold value of $\sigma$ determining whether a voxel is occupied.

% \subsection{RenderSup: 2D Supervision with Rendering}
\subsection{Rendering Supervision with 2D Labels}
\label{RenderSup}
% 基于voxel scene representation，我们使用NeRF中介绍的volumn rendering策略来渲染得到2D语义logits。具体来说，对于2D图像中的某个pixel，我们通过相机内外参计算对应的3D ray，并沿ray采样N个3D points，namely \{z_k\}^N_{k=1}。
We utilize volume rendering to form a bridge between the SDF and 2D pixels, thereby facilitating the supervision through 2D labels.
Specifically, we extract 3D rays from the current frame using camera intrinsic and extrinsic parameters, with each 2D pixel corresponding to a 3D ray originating from the camera. Each ray $\textbf{r}$ carries the semantic and depth labels $\{\hat{S}^{pix}(r),\hat{D}^{pix}(r)\}$ of the corresponding pixel. 
Meanwhile, we perform volume rendering \cite{volumerendering} based on the SDF to obtain the rendered semantic $S^{pix}(r)$ and depth $D^{pix}(r)$, which are used to compute the loss with 2D labels $\{\hat{S}^{pix}(r),\hat{D}^{pix}(r)\}$. 
% T(p) is the accumulated transmittance from the near plane to pointi;

To render the semantic and depth of a pixel, K points $\{z_k\}^K_{k=1} \in r$ are sampled on the ray $r$ in a pre-defined range. Then the accumulated transmittance $T$ and the probability of termination $\alpha$ of the point $z_k$ can be computed by
\begin{align}
\label{eq_ray_rendering}
\alpha(z_k) &= 1-exp(-\sigma(z_k)\beta_k), \\
 T(z_k) &= exp(-\sum_{t=1}^{k-1}\sigma(z_t)\beta_t), 
\end{align}
where $\beta_k=z_{k+1}-z_k$ is the distance between two adjacent points. Finally, we query the SDF with $\{z_k\}$ and accumulated them to get rendered semantic and depth:
\begin{align}
 S^{pix}(r) &= \sum_{k=1}^{N}T(z_k)\alpha(z_k)S(z_k), \\
 D^{pix}(r) &= \sum_{k=1}^{N}T(z_k)\alpha(z_k)z_k,
\end{align}

For loss fuctions, cross-entropy loss $L_{seg}$ and SILog loss $L_{depth}$ \cite{silog} are leveraged to supervise the semantic and depth, respectively. We also introduce distortion loss \cite{mipnerf360} 
 and TV loss \cite{tvloss} as the regularization of SDF, which termed $L_{reg}$.
Therefore, the overall loss can be computed by
\begin{align}
\label{loss}
 {L} =  &L_{seg}(S^{pix}, \hat{S}^{pix}) +  L_{depth}(D^{pix}, \hat{D}^{pix})
        + L_{reg}(\sigma), 
\end{align}

% 我们的方法
\begin{figure}[t]
% \vspace{-0.4cm}
\centering
\includegraphics[width=0.8\linewidth]{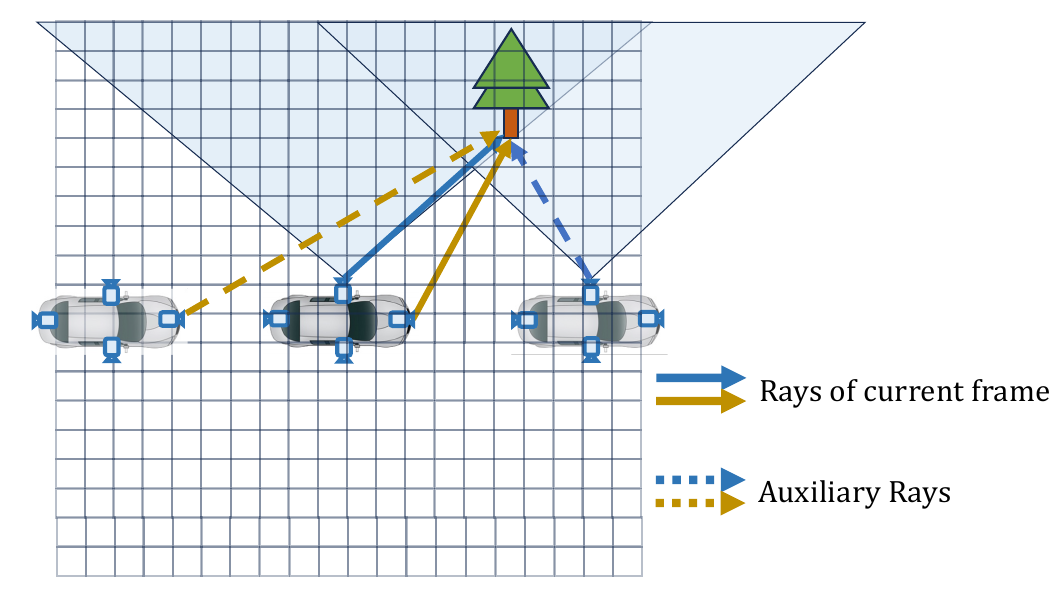}    
\vspace{-0.1cm}
\caption{\label{fig:2}
% \textbf{Auxiliary Rays Supervision Strategy.} 
\textbf{Auxiliary Rays:} Images from single frame cannot capture multi-view information of objects well. There is only a small overlap area between two adjacent cameras, and the difference in perspective is limited. By introducing auxiliary rays from adjacent frames, the model will significantly benefit from multi-view consistency constraints. 
% \textbf{(b) Ray Mask Strategy:} Due to the presence of dynamic objects, the label of auxiliary rays may be incorrect for the current frame. Therefore, we cleverly designed a mask strategy to reduce the impact of error rays.
}
\vspace{-0.2cm}
\label{fig_aux}
\end{figure}

% \subsection{Auxiliary Rays Training Strategy}
\subsection{Auxiliary Rays: Boosting Multi-view Consistency} 
\label{AuxRays}
% 借助rendering supervision，模型从多视角一致性约束中受益，学会考虑voxel之间的空间遮挡关系。然而single frame下的周视相机视角是非常稀疏的、且重叠范围有限，绝大部分voxel无法同时被多条存在较大视角差异的rays采样到，使得RenderSup容易陷入局部最优。因此，我们引入来源于时序帧的辅助射线，用于补充多视角一致性约束。
With the 2D rendering supervision in Sec. \ref{RenderSup}, the model can benefit from multi-view consistency constraints and learn to account for spatial occlusion relationships among voxels. However, the viewpoint coverage of the surrounding cameras in a single frame is very sparse, and their overlapping range is limited. As a result, most voxels cannot be simultaneously sampled by multiple rays with significant viewpoint differences, which easily leads to local optima. Therefore, we introduce auxiliary rays from adjacent frames to complement the multi-view consistency constraints as shown in Fig. \ref{fig_aux}.

\noindent\textbf{Generation of Auxiliary Rays.} 
%具体来说，对于当前帧(index=t)，我们选取临近的$N_{\text{aux}}$个时序帧。我们为为每一帧分别生成rays，并都transform到当前帧得到最终的辅助射线$r_{aux}$:
Specifically, for the current frame with index $t$, we select nearby $M_{\text{aux}}$ adjacent frames. For each adjacent frame, we generate rays individually and transform them to the current frame to obtain the final auxiliary rays $r_{aux}$:
\begin{align}
\label{aux_align}
r_{aux} = \{\mathrm{T}^{t-k}_{t}(r_{t-k}), \quad k=-1,1,\cdots, -\frac{M_{\text{aux}}}{2},\frac{M_{\text{aux}}}{2} \}
\end{align}
where $\mathrm{T}^{t-k}_{t}$ is the transformation matrix from adjacent-frame coordinates to the current frame. Given ego pose matrices $E_{t-k}$ from adjacent frame and $E_{t}$ from the current frame, we can calculate $\mathrm{T}^{t-k}_{t} = E^{inv}_t \cdot E_{t-k}$, where $E^{inv}_t$ is the inverse matrix of $E_t$.

\noindent\textbf{Weighted Ray Sampling.} 
% \subsection{Weighted Ray Sampling: Balance and Purification} 
\label{RayMask}
% NeRF：整个训练集是一个scene，所有rays被均分到data_batch中
% RenderOcc：每一帧都是一个scene，一个data_batch需要考虑这一帧所有的scene
% 辅助射线的引入可以显著提高2D监督的作用，但是随之出现了两个问题：（a）由于倍增的rays数量会导致巨大的显存和计算开销，我们不得不在训练时对rays进行随机采样，大量有价值的rays被舍弃了。（b）由于动态物体和transformation误差的存在，存在许多错误的辅助射线。To address this，我们设计了Weighted Ray Sampling策略，集中对信息密度大、相对正确的ray进行采样，在极大提高训练效率的同时还优化了性能。
The introduction of auxiliary rays significantly enhances 2D supervision but raises two challenges: (a) Increased rays lead to high memory and computational costs, necessitating random sampling during training, which discards many valuable rays. (b) Due to the presence of dynamic objects, many auxiliary rays exhibit temporal mismatches, introducing unnecessary errors. To address this, we have devised the Weighted Ray Sampling strategy, focusing on sampling high-information-density and relative correct rays. This not only significantly enhances training efficiency but also optimizes performance.
\textbf{(a) Category Density Balance:} 
% 室外自驾场景中类别不均衡现象极其严重，绝大部分rays属于路面、建筑等体积巨大、出现频率高的背景物体上，信息密度很低。相比之下行人、自行车等物体的rays是极其稀少而有价值的，这部分rays要尽可能被关注。因此，我们基于各个类别的出现频率计算$W_b$，可以被公式化为：
In outdoor autonomous driving, extreme category imbalance is common. Most rays correspond to large, low-information-density background objects like roads and buildings, while rays associated with pedestrians, bicycles are scarce but valuable.
% In outdoor self-driving scenes, extreme category imbalance is common. Most rays pertain to high-volume, low-information-density background objects like roads and buildings, while rays related to objects such as pedestrians, bicycles are rare but valuable. 
For this, we calculate weight $W_b$ based on class occurrence frequency, formulated as
\begin{align}
W_b(r) = \exp(\lambda_s*(\frac{max(M)}{N(\mathbf{C}(r))}-1))
\end{align}
where $\lambda_s$ represents a smoothing coefficient, $M$ represents the numbers of rays of all categories, and $\mathbf{C}(r)$ denotes the category of the ray $r$.
\textbf{(b) Temporal Misalignment Purification:} 
% 通过Eq. \ref{aux_align}，辅助射线可以被align到当前帧对SDF进行监督。然而，动态物体的移动会引起Misalignment，导致一部分辅助rays指向了SDF中错误的voxel。我们使用了一种简单但有效的策略来减少不匹配rays的采样概率，对动态物体进行masking并尽可能保留当前帧的射线。
By Eq. \ref{aux_align}, auxiliary rays can be aligned with the current frame for SDF supervision. However, the movement of dynamic objects can cause misalignment, leading some auxiliary rays to point to incorrect voxels in the SDF. We employed a simple yet effective strategy to reduce the sampling probability of mismatched rays, by masking dynamic objects and preserving the rays from the current frame as much as possible.  
\begin{align}
W_t(r) = \begin{cases} 
\lambda_{\text{dyn}}, &  C(r) \in C_{\text{dynamic}} \\
\lambda_{\text{adj}}, &  C(r) \notin C_{\text{dynamic}}
\end{cases}
\end{align}
where $\lambda_{\text{dyn}}$ and $\lambda_{\text{adj}}$ are coefficients less than 1 and $C_{\text{dynamic}}$ denotes the set of categories for dynamic objects. By utilizing $\lambda_{\text{adj}}$, we can lower the likelihood of rays from the current frame being omitted. Setting $\lambda_{\text{dyn}}$ to a value close to 0 allows us to significantly mitigate misalignment issues caused by dynamic objects.

% 我们为每条射线计算 $W=W_b*W_t$，作为随机采样时的概率权重。在训练时，我们通过权重W为每一个batch采样固定数量的rays。
Finally, we calculate a weight $W$ for each ray as $W = W_b \cdot W_t$. This weight acts as the probability weight of random sampling. Throughout the training process, we sample a fix number of rays for each batch using $W$, and the remaining rays are discarded and do not contribute to the loss computed by Eq. \ref{loss}.
Through proposed Weighted Ray Sampling, we can significantly reduce memory and computational costs of RenderOcc during training while achieving superior performance. Detailed discussions are provided in Sec. \ref{ablation_wrs}.

\begin{table*}[t!]
% \vspace{-3mm}
\caption{\textbf{3D occupancy prediction performance on the Occ3D-NuScenes dataset.} 
GT represents which labels we used during training. Mean is the average value of mIOU across all categories.}
\label{table_nuscenes}
\centering
\setlength\tabcolsep{2pt}%调列距
\begin{tabular}{lc|c|ccccccccccccccccc}
\hline
Method  & GT & Mean & \rotatebox{90}{Others} & \rotatebox{90}{barrier} & \rotatebox{90}{bicycle} & \rotatebox{90}{bus} & \rotatebox{90}{car} &\rotatebox{90}{cons. veh} &\rotatebox{90}{motorcycle} &\rotatebox{90}{pedestrian} &\rotatebox{90}{traffic cone} &\rotatebox{90}{trailer} &\rotatebox{90}{truck} &\rotatebox{90}{dri. sur} &\rotatebox{90}{other flat} &\rotatebox{90}{sidewalk} &\rotatebox{90}{terrain} & \rotatebox{90}{manmade}& \rotatebox{90}{vegetation} \\ \hline
MonoScene \cite{monoscene} & 3D & 6.06 & 1.75 & 7.23 & 4.26 & 4.93 & 9.38 & 5.67 & 3.98 & 3.01 & 5.90 & 4.45 & 7.17 & 14.91 & 6.32 & 7.92 & 7.43 & 1.01 & 7.65 \\
BEVFormer & 3D & 23.67 & 5.03& 38.79 & 9.98 & 34.41 & 41.09 & 13.24 & 16.50 & 18.15 & 17.83 & 18.66 & 27.70 & 48.95 & 27.73 & 29.08 & 25.38 & 15.41 & 14.46  \\
BEVStereo & 3D & 24.51 & 5.73& 38.41 &7.88 & 38.70 & 41.20 &17.56 &17.33 &14.69 &10.31 &16.84 &29.62 &54.08 &28.92 &32.68 &26.54 &18.74 &17.49 \\ 
OccFormer \cite{occformer} & 3D  & 21.93 & 5.94& 30.29 & 12.32 & 34.40 & 39.17 & 14.44 & 16.45 & 17.22 & 9.27 & 13.90 & 26.36 & 50.99 & 30.96 & 34.66 & 22.73 & 6.76 & 6.97  \\   \hline
\cellcolor{lightgray}RenderOcc (Ours) & \cellcolor{lightgray}2D & 23.93 & 5.69 &27.56 &14.36 &19.91 &20.56 &11.96 &12.42 &12.14 &14.34 &20.81 &18.94 &68.85 &33.35 &42.01 &43.94 &17.36 &22.61  \\     
RenderOcc* & 2D+3D & 26.11 & 4.84 & 31.72 & 10.72 & 27.67 & 26.45 & 13.87 & 18.2 & 17.67 & 17.84 & 21.19 & 23.25 & 63.2 & 36.42 & 46.21 & 44.26 & 19.58 & 20.72 \\  \hline

\end{tabular}
\end{table*}

\begin{table*}[t!]
% \vspace{-3mm}
\caption{\textbf{3D occupancy prediction performance on the semanticKiTTI dataset.} 
}
\label{table_kitti}
\centering
\setlength\tabcolsep{2pt}%调列距
\begin{tabular}{lc|c|ccccccccccccccccccc}
\hline
Method  & GT & Mean  & \rotatebox{90}{car} & \rotatebox{90}{bicycle} & \rotatebox{90}{motorcycle} & \rotatebox{90}{truck} & \rotatebox{90}{other-veh.} &\rotatebox{90}{person} &\rotatebox{90}{bicyclist} &\rotatebox{90}{motorcyclist} &\rotatebox{90}{road} &\rotatebox{90}{parking} &\rotatebox{90}{sidewalk} &\rotatebox{90}{other-grnd} &\rotatebox{90}{building} &\rotatebox{90}{fence} &\rotatebox{90}{vegetation} & \rotatebox{90}{trunk}& \rotatebox{90}{terrain}& \rotatebox{90}{pole}& \rotatebox{90}{traf.-sign} \\ \hline
MonoScene & 3D & 11.30 & 23.29& 0.28 & 0.59& 9.29& 2.63 & 2.00 & 1.07 & 0.00 & 55.89 & 14.75 & 26.50 & 1.63 & 13.55 & 6.60 & 17.98 & 2.44 & 29.84 & 3.91 & 2.43 \\
LMSCNet \cite{lmscnet} & 3D & 9.94& 23.62 & 0.00 & 0.00 & 1.69 & 0.00 & 0.00 & 0.00 & 0.00 & 54.90 & 9.89 & 25.43 & 0.00 & 14.55 & 3.27 & 20.19 & 1.06 & 32.30 & 2.04 & 0.00  \\
AICNet \cite{aicnet} & 3D & 6.73 & 15.30 & 0.00 & 0.00 & 0.70 & 0.00 & 0.00 & 0.00 & 0.00 & 39.30 & 19.80 & 18.30 & 1.60 & 9.60 & 5.00 & 9.60 & 1.90 & 13.50 & 0.10 & 0.00  \\
VoxFormer \cite{voxformer} & 3D & 12.35 & 25.79 & 0.59 & 0.51 & 5.63 & 3.77 & 1.78 & 3.32 & 0.00 & 54.76 & 15.50 & 26.35 & 0.70 & 17.65 & 7.64 & 24.39 & 5.08 & 29.96 & 7.11 & 4.18  \\  \hline
\cellcolor{lightgray}RenderOcc (Ours) & \cellcolor{lightgray}2D & 8.24 & 14.83 & 0.42 & 0.17 & 2.47 & 1.78 & 0.94 & 3.20 & 0.00 & 43.64 & 12.54 & 19.10 & 0.00 & 11.59 & 4.71 & 17.61 & 1.48 & 20.01 & 1.17 & 0.88  \\   
RenderOcc* & 2D+3D & 12.87 &24.90 & 0.37 & 0.28 & 6.03 & 3.66 & 1.91 & 3.11 & 0.00 & 57.2 & 16.11 & 28.44 & 0.91 & 18.18 & 9.10 & 26.23 & 4.87 & 33.61 & 6.24 & 3.38   \\     \hline

\end{tabular}
\end{table*}

\section{Experiments}
% We evaluate RenderOcc for visual-centric 3D occupancy prediction task. 
% % Our method is universal and can support different camera settings and network structures. 
% In Sec. \ref{main}, 
We evaluate proposed RenderOcc by comparing it with other baselines on NuScenes and SemanticKiTTI.
% We conducted evaluations in both NuScenes and SemanticKiTTI scenarios. 
For a deeper understanding of RenderOcc, we also conducted extensive ablation experiments on NuScenes dataset in Sec. \ref{ablation}.
% In Sec~\ref{sec:exp1}, we provide the details of the task settings for vision-centric 3D occupancy prediction, as well as a description of the datasets \cite{nuscenes, semantickitti}. In Sec~\ref{sec:4.2}, we evaluate our method by comparing it with other baselines ~\cite{monoscene, lmscnet, aicnet,bevformer, bevstereo, occformer}. Comprehensive ablation studies on NuScenes are conducted in Sec~\ref{sec:4.3},For a deeper understanding of RenderOcc, we conducted extensive ablation experiments on NuScenes.
% Comprehensive ablation studies are conducted in Sec~\ref{sec:4.3}, which investigate the impact of each component.

\subsection{Dataset} 
\label{sec:exp1}
We evaluate our method on NuScenes and SemanticKiTTI, respectively for surrounding-view and front-view setting. The NuScenes dataset includes 1000 outdoor driving scenes with six surrounding-view cameras, and the 3D occupancy ground truth provided by \cite{occ3d} of each sample covers a range of [-40m, -40m, -1m, 40m, 40m, 5.4m] with a voxel size of [0.4m,0.4m,0.4m]. It has 17 classes. 
% For NuScenes, we use methods such as MonoScene, BEVFormer, VoxFormer, BEVStereo, etc. that supportting multi-view cameras as baselines, and use same setting like BEVStereo to evaluate RenderOcc. 
The SemanticKiTTI dataset includes 22 outdoor driving scenes, and the benchmark is interested in areas ahead of the car. Each sample covers a range of [0.0m, -25.6m, -2.0m, 51.2m, 25.6m, 4.4m] with a voxel size of [0.2m,0.2m,0.2m]. It has 19 classes. 
% For SemanticKiTTI, we use methods such as MonoScene, AICNet, LMSCNet, VoxFormer, etc. as baselines, and use same setting like VoxFormer to evaluate RenderOcc.
Neither NuScenes nor SemanticKITTI directly provide ground truth (GT) for 2D segmentation and depth. Therefore, we project 3D LiDAR points with segmentation labels onto images to generate 2D labels.

\subsection{Architecture and Implementation Details}
% RenderOcc对于$G$(Volumn feature encoder)的网络架构并不敏感。
% RenderOcc is insensitive to the architecture of $\mathbb{G}$ (2D-to-3D encoder that extract volumn features from multi-camera images).
We use the available network BEVStereo \cite{bevstereo} as $\mathbb{G}$ to test the performance of RenderOcc. We maintain the majority of the original structure and only replace the classification head $\mathbb{F}$ with $\{\phi_d, \phi_s\}$ to predict the Semantic Density Field, thereby supporting the training by 2D rendering supervision.
For $\mathbb{G}$, we use Swin Transformer \cite{swin} as the image backbone, resize the image to 512x1408, use Adam as the optimizer, set the Batchsize to 16, and train approximately $\sim 10K$ iters with a learning rate of 1e-4. All experiments are conducted on NVIDIA A100 GPUs.

\subsection{Main Results} 
\label{main}
\begin{figure}[t]
% \vspace{0.1cm}
\centering
\includegraphics[width=1.0\linewidth]{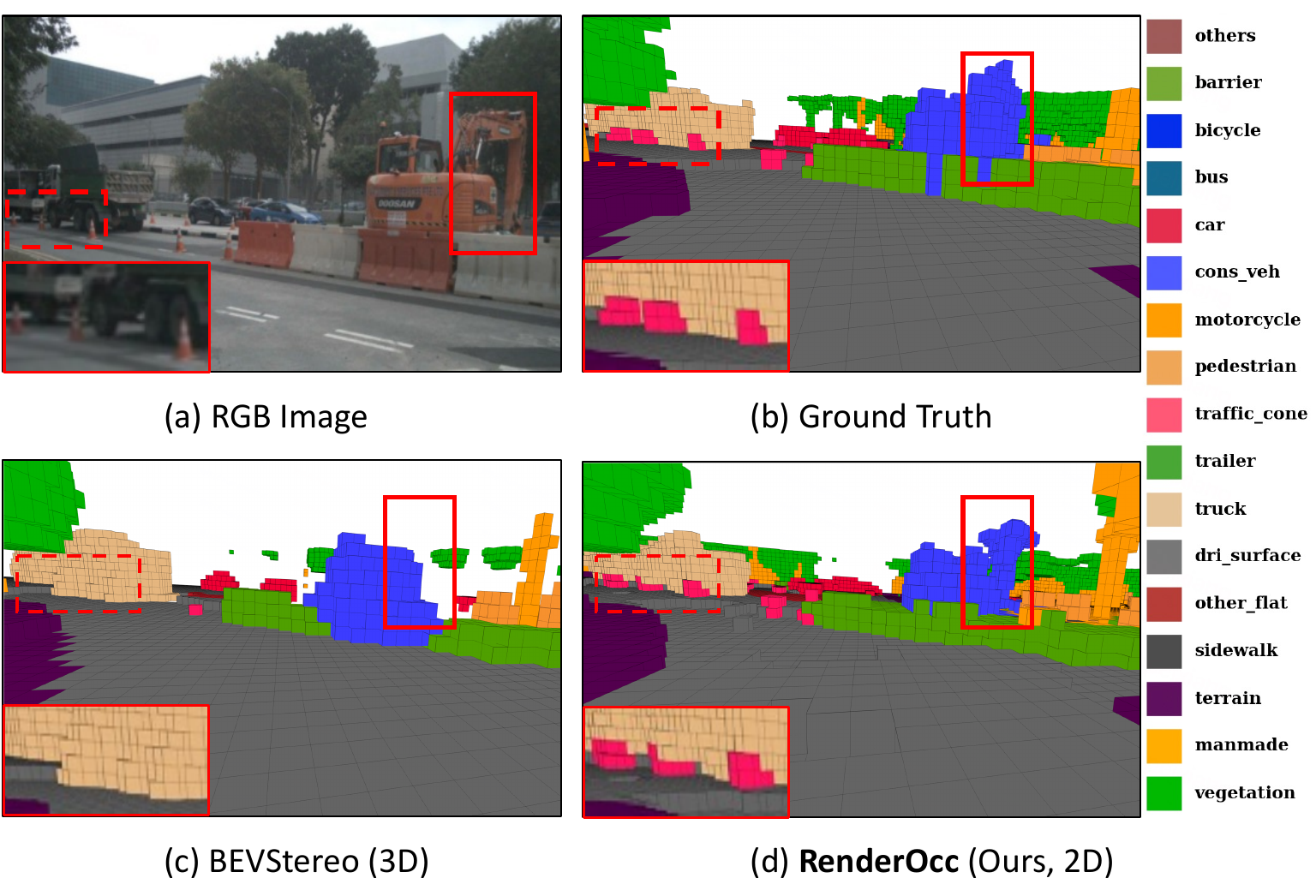}    
\vspace{-0.3cm}
\caption{\label{fig:1}
% \textbf{Auxiliary Rays Supervision Strategy.} 
\textbf{Qualitative results on NuScenes.} 
Compared to the baseline that uses 3D labels for supervision, our proposed RenderOcc exhibits a more acute perception of object boundaries and small objects as shown in the red boxes. The crane’s arm in the image is finely perceived by RenderOcc, while BEVStereo supervised by 3D labels fails to perceive the arm floating in the air. At the same time, RenderOcc successfully identifies distant traffic cones that the baseline overlooks. 
}
\vspace{-0.4cm}
\label{fig_vis}
\end{figure}

\noindent\textbf{Results on NuScenes.} 
In this section, we provide experimental results on the NuScenes dataset, as shown in Tab. \ref{table_nuscenes}.
We employs 6 adjacent frames to generate auxiliary rays and sample 38400 rays for a batch.
RenderOcc, when supervised solely with 2D labels, achieves a mean mIoU of 23.93, demonstrating competitive results when compared to Baselines supervised by 3D occupancy labels. 
In details, RenderOcc exhibits only a marginal decline of 0.58 mIoU compared with BEVStereo, surpassing other baselines such as MonoScene, OccFormer and BEVFormer. Under our proposed 2d rendering supervision, RenderOcc performs significant variations in its performance across different categories compared to 3D supervised baselines.
For static background categories like driveable surfaces, terrain, and vegetation, RenderOcc achieves exceptionally high mIoU, effectively identifying road structures, benefiting from an explicit understanding of 3D spatial relationships. 
However, RenderOcc's performance is relatively poorer for dynamic objects, often predicting artifacts such as ghosting. Our proposed Auxiliary Rays strategy can alleviate this issue to some extent.
For small foreground objects like bicycles, motorcycles, and traffic cones, RenderOcc also outperforms Baselines, thanks to the fine-grained supervision provided by 2D pixel-level labels. Additionally, RenderOcc supports simultaneous supervision using both 2D and 3D labels, achieving the best result with an mIoU of 26.11. The result demonstrates that our method of 2D rendering supervision can aid in improving the construction of voxel representations for existing 3D occupancy labels and enhance overall understanding of 3D scenes. Finally, we conducted a qualitative analysis in Fig .\ref{fig_vis}, and observed that our method yields improved accuracy in capturing semantic information and the shapes of objects within the 3D scene.

% 我们的方法

\noindent\textbf{Results on SemanticKITTI.} 
% In this section, we provide experimental results on the SemanticKITTI dataset, as shown in Tab. \ref{table_kitti}。我们模型的特征提取网络使用VoxFormer，辅助射线的帧数设置为5、rays数量上限设为25600。可以看到，与使用3D Occupancy Label监督的方法相比，仅使用2D标签监督的RenderOcc能够取得9.44的mIoU，超过部分Baseline。不幸的是，SemanticKITTI仅使用单个前视相机，相对固定且单一的视角导致了多视角一致性约束的缺乏，导致RenderSup陷入局部最优。辅助射线能够一定程度上缓解此问题，但最终相比MonoScene和VoxFormer仍旧存在2～3 mIoU的差距。在此基础上加入已有的3D标签，RenderOcc能够客服单一视角的优化问题，取得3.4的mIoU提升，这证明了RenderOcc的适应性。
% In this section, we provide experimental results on the SemanticKITTI dataset, as shown in Tab. \ref{table_kitti}.
% We employs 4 adjacent frames to generate auxiliary rays and sample 25600 rays for a batch.
% RenderOcc achieves an mIoU of 8.24 supervised solely using 2D label, surpassing some of the baseline methods that supervised by 3D occupancy GT. Unfortunately, SemanticKITTI relies on a single front-facing camera, leading to a lack of multi-view consistency constraints due to its fixed and singular viewpoint. This limitation causes RenderSup to become trapped in local optima. While auxiliary rays partially mitigate this issue, there remains a 2-3 mIoU gap compared to MonoScene and VoxFormer.
% Furthermore, when incorporating existing 3D labels, RenderOcc overcomes the optimization problem stemming from the single viewpoint, resulting in a 12.87 mIoU. This underscores RenderOcc's adaptability.
In this section, we provide experimental results on the SemanticKITTI dataset, as shown in Tab. \ref{table_kitti}.
We employs 4 adjacent frames to generate auxiliary rays and sample 25600 rays for a batch.
RenderOcc achieves a mean mIoU of 8.24 supervised solely using 2D label, showing competitive result with other baseline methods that supervised by 3D occupancy GT. Unfortunately, SemanticKITTI relies on a single front-facing camera, leading to a lack of multi-view consistency constraints due to its fixed and singular viewpoint. This limitation causes 2D rendering supervision to become trapped in local optima.
While auxiliary rays strategy partially mitigate this issue, it validates the effectiveness of our approach in monocular 3D occupancy prediction.
Furthermore, when incorporating existing 3D labels, RenderOcc overcomes the optimization problem stemming from the single viewpoint, resulting in a 12.87 mIoU. This result showcases the practicality of our method in real-world applications, whether the input images are from multi-view cameras or single-view camera.

\begin{table}[t!]
\caption{\textbf{Ablation study for each component on the Occ3D-NuScenes.}}
\label{table_ablation}
\centering
\begin{tabular}{cccc|c}
\hline
RenSup-S & RenSup-D & Aux Rays & WRS & \cellcolor{lightgray} mIoU ↑\\ \hline
\checkmark &  &  & & 16.94  \\
\checkmark & \checkmark &  & & 19.28 \\
\checkmark & \checkmark & \checkmark & & 22.41 \\
\checkmark & \checkmark & \checkmark &\checkmark & \textbf{23.93} \\ \hline
\end{tabular}
\end{table}

\subsection{Ablation Study \& Analysis} 
\label{ablation}
% To better reflect the role of each component in RenderOcc, we conduct ablation experiments to analyze how each component benefit. 

\begin{table}[t!]
% \caption{\textbf{Ablation study for rendering sampling ways.}
% The choice of sampling methods during training can have a certain impact on model performance.
% }
\caption{\textbf{Ablation study for rendering sampling ways.} 
% A smaller “step size” represents denser sampling of 3D points on the ray.
% "Step size" represents the smallest unit of a voxel, reflecting the voxel's resolution. 
% "Point number" represents the quantity of rays in each image. 
}
\label{table_sampleing}
\centering
\begin{tabular}{c|cc|c}
\hline
Method & step size & point number & \cellcolor{lightgray} mIoU ↑\\ \hline
unified-sampling & 1.0*voxel\_size & / & 19.42  \\
unified-sampling & 0.5*voxel\_size & / & 21.10  \\ \hline
hierarchical-sampling & / & 64$\rightarrow$64 & 22.34  \\  
hierarchical-sampling & / & 64$\rightarrow$128 & 22.98  \\   \hline
% mip360-sampling(4f) & 1.0*voxel\_size& / & 22.49  \\ 
% mip360-sampling(4f) & 0.5*voxel\_size& / & 23.44  \\ \hline
mip360-sampling & 1.0*voxel\_size& / & 22.61  \\ 
mip360-sampling & 0.5*voxel\_size& / & \textbf{23.93}  \\ \hline     % transfer to 6 aux frames
\end{tabular}
\end{table}

\begin{table}[t!]
      \begin{center}
        \caption{\textbf{Analysis of depth supervision.}
        "LiDAR projection" refers extracting depth labels from LiDAR point clouds. "Struct From Motion" represents deriving depth labels from image-based algorithms \cite{sfm} without LiDAR data.
            % Utilizing depth supervision can significantly boost the performance of RenderOcc, even when the depth labels are not derived from LiDAR point clouds, but are generated based on image algorithms.
        }
        % \vspace{-0.2cm}
        \setlength{\tabcolsep}{1.2mm}{
       	\begin{tabular}{cc|c}
       	\hline
     Depth Label & Label Modality & \cellcolor{lightgray} mAP ↑ \\
        \hline
        \hline
        / & / & 16.92 \\
        LiDAR Projection & L & \textbf{23.93} \\
        Struct From Motion & C &21.11 \\   % 这里是4f的SFM结果，擦
        \hline
		\end{tabular}}
      \label{table_depth}
      \vspace{-0.2cm}
      \end{center}
\end{table}

% \begin{table}[t!]
% \caption{\textbf{Ablations on the usage of Auxiliary Rays} 
% The performance of RenderOcc increases with the number of auxiliary frames, but gradually converges when it reaches 5 frames. Introducing the Mask strategy at this point can reduce the impact of error rays and further improve model performance.)
% }
% \label{table_auxray}
% \centering
% % \setlength\tabcolsep{5pt}%调列距
% \begin{tabular}{cc|cc|c}
% \hline
% Aux frame & Ray Sampling & Ray number & Memory & \cellcolor{lightgray} mIoU ↑\\ \hline
% - & - & \~19200 & 12GB & 19.28 \\     \hline
% 2 & - & \~57600 & 15GB & 22.19  \\ 
% 4 & - & \~96000 & 24GB & 22.89  \\    
% 6 & - & \~134400 & 28GB &  23.01 \\    \hline
% 2 & \checkmark & 38400 & 13GB & 22.17  \\  
% 4 & \checkmark & 38400 & 13GB & 23.44 \\  
% 6 & \checkmark & 38400 & 13GB & 23.93 \\  \hline
% \end{tabular}
% \end{table}

\noindent\textbf{Ablation Study for each Component}
% To better reflect the role of each component in RenderOcc, 我们在Tab. \ref{table_ablation}中展示了ablation result来展示每个模块的贡献。在仅使用2D语义标签进行监督时(RenSup-S)，RenderOcc能取得16.94 mIoU，在此基础上加入Depth监督能够得到2.34 mIoU的提升(RenSup-D)。辅助射线的引入能够进一步带来2.91的mIoU提升，但代价是更高的训练开销(Aux Rays)。使用提出的Weighted Ray Sampling 策略，对Rays采样概率进行两种形式的加权，总共能带来1.25的mIoU提升（$WRS-W_b$ and $WRS-W_t)$. 更为重要的是，WRS可以解决Aux Rays带来的训练开销问题，在下文我们会进行详细讨论。
To clarify RenderOcc's component contributions, we present ablation results in Table \ref{table_ablation}. When supervised solely with 2D semantic labels (RenSup-S), RenderOcc achieves an mIoU of 16.94. Adding depth supervision (RenSup-D) increases mIoU by 2.34. Introducing the auxiliary rays strategy (Aux Rays) further boosts performance by 3.13 mIoU but at the cost of increased training overhead. Utilizing the Weighted Ray Sampling strategy yields a total mIoU improvement of 1.52. Importantly, WRS effectively addresses the training overhead introduced by auxiliary rays, which we will discuss in detail later.

\noindent\textbf{Selection for Points Sampling Ways}
% 我们对比了几种典型的Point Sampling方式，分别为unified, coarse-to-fine和NDC introduced in \cite{nerf++}. 正如Tab. \ref{table_sampling}中所示，coarse-to-fine和NDC这两种方式显著优于unified Sampling，且随着采样频率的增加能够有明显的性能提升。这可能是因为后两者能够帮助模型更好的应对large-scale且unbounded的场景，实现更加精细的采样。
We compared three typical point sampling methods for $z_k$ and make corresponding adjustments to the rendering formula., namely unified-sampling (perform sampling with a fixed step size) , hierarchical-sampling \cite{nerf} and mip360-sampling (introduced in mip-NeRF 360 \cite{mipnerf360} to address unbounded scenes). As shown in Table \ref{table_sampleing}, both hierarchical-sampling and MN360-sampling methods outperform unified-sampling.  
Moreover, with an increase in sampling frequency (step size), there is a noticeable improvement in performance. Therefore, we finally select mip360-sampling for volumn rendering.
% This may be attributed to the fact that the latter two methods assist the model in better handling large-scale and unbounded scenes, enabling more precise sampling and finally benefits occupancy prediction.

\begin{figure}[t]
\includegraphics[width=1.0\linewidth]{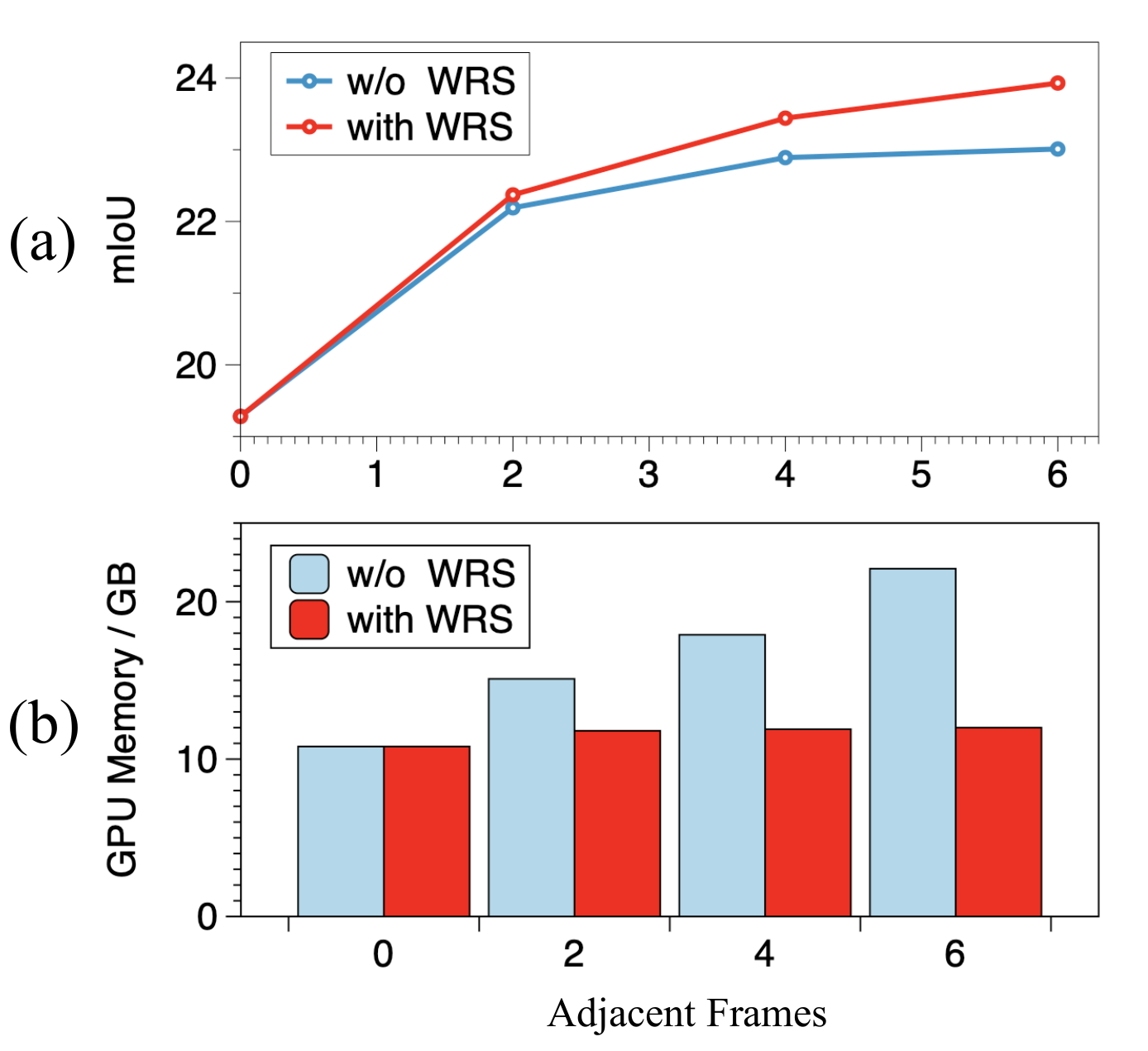}
\centering
\vspace{-0.3cm}
\caption{
% \textbf{Ablation study for Auxiliary-Ray and Weighted-Ray-Sampling.}
% As the usage of adjacent frames grows, so does the mIoU. Weighted Ray Sampling (WRS) can futhormore avoid auxiliary rays’ extra training cost while enhancing performance.
\textbf{Ablation Study For Auxiliary-Ray.}
(a) With the increased utilization of adjacent frames, there is a corresponding rise in mIoU. 
(b) Weighted Ray Sampling (WRS) effectively mitigates the additional training cost associated with auxiliary rays while improving performance.
}
\label{fig:AR}
\vspace{-0.3cm}
\end{figure}

\noindent\textbf{Auxiliary Rays and Weighted Ray Sampling}
\label{ablation_wrs}
% 辅助射线能够带来显著的收益，但是弊端是明显的：在训练时消耗更多的显存和计算开销，并且不可避免的存在misaligned rays。我们对其做了详细的消融实验，结果如Table. \ref{table_auxray}中所示。在不使用Weighted Ray Sampling的情况下,当辅助帧的数量从0帧提升到6帧时，mIoU从19.28逐渐提升至23.01，但是显存占用会随着Rays number的倍增而显著增加，产生额外的训练成本。引入Weighted Ray Sampling后，我们将Ray numbers固定在38400，集中对高信息密度的有价值Rays进行采样，在不引起额外开销的情况下取得更高的mIoU，在Aux frames=6时能取得xxx的mIoU。
As shown in Table. \ref{table_ablation}, the usage of auxiliary rays can yield significant benefits, but there are clear drawbacks: they consume more GPU memory and computational resources during training and inevitably introduce misaligned rays. We conducted comprehensive ablation experiments, and the results are presented in Fig .\ref{fig:AR}. Without using Weighted Ray Sampling, as the number of auxiliary frames increases from 0 to 6, the mIoU gradually improves from 19.28 to 23.01. However, as shown in Fig. \ref{fig:AR}(b), GPU memory usage increases significantly with the doubling of ray numbers, incurring additional training costs. With the introduction of Weighted Ray Sampling, we keep the Ray numbers fixed at 38400 and focus on sampling valuable rays with high information density, achieving higher mIoU without incurring additional overhead. 
% With 6 adjacent frames used, we attain an mIoU of 23.93 with negligible GPU memory overhead. 

\noindent\textbf{Training Without LiDAR}
In Table \ref{table_depth}, we compare various depth supervision methods. Without depth supervision, RenderOcc scores 16.9 mIoU. 
Leveraging Raw LiDAR for depth supervision yields an mIoU of 23.44. However, expensive LiDAR data is often inaccessible in many scenarios. Therefore, we also attempted to compute depth labels directly from images using a struct-from-motion approach, following \cite{surrounddepth}. Although the generated depth labels are quite sparse, RenderOcc still achieves 21.11 mIoU. This demonstrates potential in training 3D Occupancy models without LiDAR data, meriting deeper investigation in future work.

\section{Conclusion}
% In this paper, we propose RenderOcc，一个novel paradigm for training multi-camera 3D occupancy models using 2D labels.使用2D标签来训练multi-camera 3D occupancy模型的工作。 

We dive into the potential of using 2D image labels to train 3D occupancy networks, and propose a general framework RenderOcc to effectively implement this idea. This approach circumvents the production of costly and ambiguous 3D occupancy labels, training vision-centric models in a cheap and more intuitively direct manner. The extensive experiments validate the effectiveness of our method, offering a new perspective for the community.
% We dive into the potential of 使用2D图像标签来训练multi-camera 3D occupancy网络，并提出了一个general framework RenderOcc to 实现 the idea. 这种做法规避了costly and ambiguous的3D occupancy labels的制作。The experiments valiadate the effectiveness if out method，为社区提供了一种新的思路。

\section{Acknowledgement}
This research was partly supported by the foundation of the National Key R\&D Program of China (2022ZD0116305).
\clearpage
{
\bibliographystyle{IEEEtran}
\bibliography{IEEEabrv,reference}
}

\end{document}